\documentclass[letter]{article}

\newif\ifarXiv
\arXivtrue 

\PassOptionsToPackage{numbers, sort&compress}{natbib}
\usepackage{nccmath}

\ifarXiv
    \usepackage[preprint]{neurips_2021}
\else
    \usepackage{neurips_2021}
    
\fi

\usepackage[utf8]{inputenc} 
\usepackage[T1]{fontenc}    
\usepackage{hyperref}       
\usepackage{url}            
\usepackage{booktabs}       
\usepackage{amsfonts}       
\usepackage{nicefrac}       
\usepackage{microtype}      
\usepackage{xcolor}         

\usepackage{amsmath,amsfonts,bm}

\def\eqref#1{equation~\ref{#1}}

\def\1{\bm{1}}

\def\vmu{{\bm{\mu}}}
\def\vsigma{{\bm{\sigma}}}
\def\vtheta{{\bm{\theta}}}
\def\vomega{{\bm{\omega}}}
\def\vphi{{\bm{\phi}}}

\def\ve{{\bm{e}}}

\def\vh{{\bm{h}}}

\def\vm{{\bm{m}}}

\def\vs{{\bm{s}}}

\def\vv{{\bm{v}}}

\def\vx{{\bm{x}}}

\def\vz{{\bm{z}}}

\def\ftau{\fvec{\tau}}

\def\vfmu{\fvec{\vmu}}
\def\vfsigma{\fvec{\vsigma}}

\def\vfomega{\fvec{\vomega}}

\def\vfh{\fvec{\vh}}

\def\vfx{\fvec{\vx}}

\def\vfz{\fvec{\vz}}

\def\btau{\cev{\tau}}

\def\vbmu{\cev{\vmu}}
\def\vbsigma{\cev{\vsigma}}

\def\vbomega{\cev{\vomega}}

\def\vbh{\cev{\vh}}

\def\vbx{\cev{\vx}}

\def\vbz{\cev{\vz}}

\DeclareMathAlphabet{\mathsfit}{\encodingdefault}{\sfdefault}{m}{sl}
\SetMathAlphabet{\mathsfit}{bold}{\encodingdefault}{\sfdefault}{bx}{n}

\def\sI{{\mathbb{I}}}

\def\sT{{\mathbb{T}}}

\newcommand{\E}{\mathbb{E}}
\newcommand{\Ls}{\mathcal{L}}
\newcommand{\Ltwo}{\mathcal{L}_{2}}
\newcommand{\R}{\mathbb{R}}

\usepackage{comment}
\usepackage{graphicx,accents}
\usepackage{xspace}
\usepackage{xcolor}
\usepackage[subrefformat=parens,labelformat=parens]{subcaption}
\usepackage{booktabs}
\usepackage{bibunits}
\usepackage{siunitx}
\usepackage{appendix}
\usepackage{pifont}

\usepackage{pdflscape}
\usepackage{afterpage}

\newcommand{\cmark}{\ding{51}}%
\newcommand{\xmark}{\ding{55}}%

\newcommand{\bi}{\begin{itemize}}
\newcommand{\ei}{\end{itemize}}
\newcommand{\be}{\begin{enumerate}}
\newcommand{\ee}{\end{enumerate}}
\newcommand{\bb}{\begin{block}}
\newcommand{\eb}{\end{block}}

\usepackage{tabularx}
\usepackage{algorithm}
\usepackage[noend]{algpseudocode}
\usepackage{algorithmicx}
\makeatletter
\newcommand{\multiline}[1]{%
  \begin{tabularx}{\dimexpr\linewidth-\ALG@thistlm}[t]{@{}X@{}}
    #1
  \end{tabularx}
}
\makeatother

\usepackage[capitalise]{cleveref}  
\crefname{equation}{}{}

\crefname{supp}{Supplementary Material}{Supplementary Materials}

\makeatletter
\DeclareRobustCommand{\cev}[1]{%
  {\mathpalette\do@cev{#1}}%
}
\newcommand{\do@cev}[2]{%
  \vbox{\offinterlineskip
    \sbox\z@{$\m@th#1 x$}%
    \ialign{##\cr
      \hidewidth\reflectbox{$\m@th#1\vec{}\mkern4mu$}\hidewidth\cr
      \noalign{\kern-\ht\z@}
      $\m@th#1#2$\cr
    }%
  }%
}
\makeatother

\makeatletter
\DeclareRobustCommand{\fvec}[1]{%
  {\mathpalette\do@fvec{#1}}%
}
\newcommand{\do@fvec}[2]{%
  \vbox{\offinterlineskip
    \sbox\z@{$\m@th#1 x$}%
    \ialign{##\cr
      \hidewidth{$\m@th#1\vec{}\mkern4mu$}\hidewidth\cr
      \noalign{\kern-\ht\z@}
      $\m@th#1#2$\cr
    }%
  }%
}
\makeatother

\newcommand\mathtight
{%
    \thinmuskip=1mu
    \medmuskip=1mu
    \thickmuskip=1mu
}
\def\LSTM{\textsc{LSTM}}
\def\GN{\textsc{GN}}

\def\pri{\text{pri}}
\def\enc{\text{enc}}
\def\dec{\text{dec}}

\newcolumntype{P}[1]{>{\centering\arraybackslash}p{#1}}

\newcolumntype{Y}{>{\centering\arraybackslash}X}

\newcounter{row}

\newlength{\tempdima}
\newcommand{\rowname}[1]{
    \rotatebox{90}{\makebox[\tempdima][c]{#1}}
}

\def\gameSubfigWidth{0.33\textwidth}
\newcommand{\resultsRow}[3]{ 
    \settoheight{\tempdima}{\includegraphics[width=\gameSubfigWidth]{figs/#1_example_0.png}}%
    \rowname{#2}&
    \includegraphics[width=\gameSubfigWidth]{figs/#1_example_#3.png}&
    \includegraphics[width=\gameSubfigWidth]{figs/#1_example_\the\numexpr#3+1.png}&
    \includegraphics[width=\gameSubfigWidth]{figs/#1_example_\the\numexpr#3+2.png}\\
    \stepcounter{row}%
}

\newcommand{\gc}[1]{\textbf{\color{red}[#1]}}
\newcommand{\so}[1]{\textbf{\color{red}[SO: #1]}}
\newcommand{\re}[1]{\textbf{\color{purple}[$\rho\mu$: #1]}}
\newcommand{\kt}[1]{\textbf{\color{purple}[KT: #1]}}
\newcommand{\ma}[1]{\textbf{\color{magenta}[MG: #1]}}
\renewcommand{\dh}[1]{\textbf{\color{teal}[DH: #1]}}

\renewcommand{\gc}[1]{\ignorespaces}
\renewcommand{\so}[1]{\ignorespaces}
\renewcommand{\re}[1]{\ignorespaces}
\renewcommand{\kt}[1]{\ignorespaces}
\renewcommand{\ma}[1]{\ignorespaces}
\renewcommand{\dh}[1]{\ignorespaces} 

\usepackage{selectp}

\newcommand{\papertitle}{Time-series Imputation of Temporally-occluded Multiagent Trajectories}

\title{\papertitle}

\author{%
    Shayegan Omidshafiei$^{1}$\\
    {\small \texttt{somidshafiei@deepmind.com}} \\
    \And 
	Daniel Hennes$^{1}$\\
	{\small \texttt{hennes@deepmind.com}}\\
	\And
	Marta Garnelo$^{1}$\\
	{\small \texttt{garnelo@deepmind.com}}\\
	\And
	Eugene Tarassov$^{1}$\\
	{\small \texttt{etar@deepmind.com}}\\
	\And
	Zhe Wang$^{1}$\\
	{\small \texttt{zhewang@deepmind.com}}\\
	\And
	Romuald Elie$^{1}$\\
	{\small \texttt{relie@deepmind.com}}\\
	\And
	Jerome T. Connor$^{1}$\\
	{\small \texttt{jeromeconnor@deepmind.com}}\\
	\And
	Paul Muller$^{1}$\\
	{\small \texttt{pmuller@deepmind.com}}\\
	\And
	Ian Graham$^{2}$\\
	{\small \texttt{ian.graham@liverpoolfc.com}}\\
	\And
	William Spearman$^{2}$\\
	{\small \texttt{william.spearman@liverpoolfc.com}}\\
	\And
    Karl Tuyls$^{1}$\\
    {\small \texttt{karltuyls@deepmind.com}} \\
    \AND
    {\normalfont $^1$DeepMind $\hspace{5pt}$ $^2$Liverpool Football Club} \\
}

\begin{document}

\maketitle

\begin{abstract}
In multiagent environments, several decision-making individuals interact while adhering to the dynamics constraints imposed by the environment. These interactions, combined with the potential stochasticity of the agents' decision-making processes, make such systems complex and interesting to study from a dynamical perspective. Significant research has been conducted on learning models for forward-direction estimation of agent behaviors, for example, pedestrian predictions used for collision-avoidance in self-driving cars. However, in many settings, only sporadic observations of agents may be available in a given trajectory sequence. For instance, in football, subsets of players may come in and out of view of broadcast video footage, while unobserved players continue to interact off-screen. In this paper, we study the problem of multiagent time-series imputation, where available past and future observations of subsets of agents are used to estimate missing observations for other agents. Our approach, called the Graph Imputer, uses forward- and backward-information in combination with graph networks and variational autoencoders to enable learning of a distribution of imputed trajectories. We evaluate our approach on a dataset of football matches, using a projective camera module to train and evaluate our model for the off-screen player state estimation setting. We illustrate that our method outperforms several state-of-the-art approaches, including those hand-crafted for football.
\end{abstract}

\section{Introduction}\label{sec:introduction}

Predictive modeling of multiagent behaviors has been a topic of considerable interest in machine learning \citep{FoersterFANW18,BrownLGS19,sun2018predicting}, financial economics \citep{Taylor07,Sezer20,tay2001application}, robotics \citep{Alahi_2016_CVPR,Sakata18,Rudenko20}, and sports analytics \citep{le2017coordinated,le2017data,yeh2019diverse,hauri2020multimodal}. 
In such systems, decision-making agents interact within a shared environment, following an underlying dynamical process that may be stochastic and, at times, infeasible to characterize analytically due to the complex interactions involved.
Learning a dynamical model of such systems, importantly, enables both the understanding and evaluation of agents' behaviors.
Ideally, methods that learn models of such coupled dynamical systems should enable the prediction of future behaviors, the retrodiction of past behaviors, and ultimately the imputation (i.e., filling-in) of partially-occluded data, while adhering to any constraints imposed by available observations. 
In this paper, we introduce such a method for multiagent time-series imputation under temporal occlusion.
We evaluate our method in the football domain, where trajectory prediction of off-screen players is of interest as it can be used for counterfactual reasoning for coaching purposes and within downstream football analytics models that require fully observed data (e.g., pitch control~\citep{spearman2017physics}).

Football is an especially interesting testbed for the multiagent imputation problem as it involves mixed cooperative and competitive interactions between players, is stochastic due to the human decisions involved, and includes players whose roles can dynamically change throughout the match (e.g., defenders that may exhibit attacking behavior, and vice versa).
A large corpus of prior works have targeted learning models for forward-prediction of multiagent trajectories~\citep{hoshen2017,le2017coordinated,le2017data,yeh2019diverse,alcorn2021baller2vec,hauri2020multimodal,SuHSP17,Suda19,li2020Evolvegraph}.
In these works, a stream of observations for all involved entities (e.g., all players and the ball) is assumed to be available for some number of timesteps, after which the states of a subset of entities are predicted.
However, tracking data is typically sourced from third-party providers (e.g., \citet{SecondSpectrum} and \citet{Tracab}) and is not available for all matches due to the cost of proprietary sensors involved.
By contrast, video footage is widely-available for matches and can be used to track on-screen players, though does not provide explicit information about off-screen players.
In contrast to prior works, we target the under-explored multiagent imputation regime detailed above, wherein we assume available observations of on-screen players (e.g., as obtained from a vision-based tracking system), and seek to predict the unobserved states of off-screen players. 

Imputation of multivariate time series data involving interacting entities has various practical applications besides football.
In financial markets, certain foreign exchange quotations are available more frequently than others, yet correlations between these financial instruments can be used to impute the missing data~\citep{Taylor07,Sezer20,moritz2017imputets}.
In clinical trials, multi-sensory data may be made with irregular measurements or unavailable for some sensors at certain times~\citep{silva2012predicting}.
In natural language processing, in-filling of text conditioned on surrounding sentence context is an area of active research~\citep{fedus2018maskgan}, and can naturally extend to multiagent conversational dialogue in-filling.
While we are primarily motivated by the football domain, the method introduced in this paper applies to a number of multiagent settings as it makes few assumptions about the underlying dynamics. 

The contribution of this paper is a technique for bidirectional multiagent data-imputation, which permits dynamic occlusion of any subset of agents in a given trajectory sequence, and can additionally be used for forward- and backward-prediction rollouts seamlessly.
Our model uses a combination of bidirectional variational LSTMs~\citep{chung2015} and graph networks~\citep{battaglia2018}, with elements in place to handle arbitrarily-complex occlusions of sensory observations in multiagent settings.
Our experiments are conducted on a large suite of 105 full-length real-world football matches, wherein we compare our method against a number of existing approaches including Social LSTMs~\citep{Alahi_2016_CVPR} and graph variational RNNs (GVRNNs)~\citep{yeh2019diverse,sun2018predicting}. 

\section{Problem Formulation}\label{sec:prelims}

\begin{figure}[t]
    \centering
    \begin{subfigure}{0.35\textwidth}
        \centering
        \includegraphics[page=1,width=\textwidth]{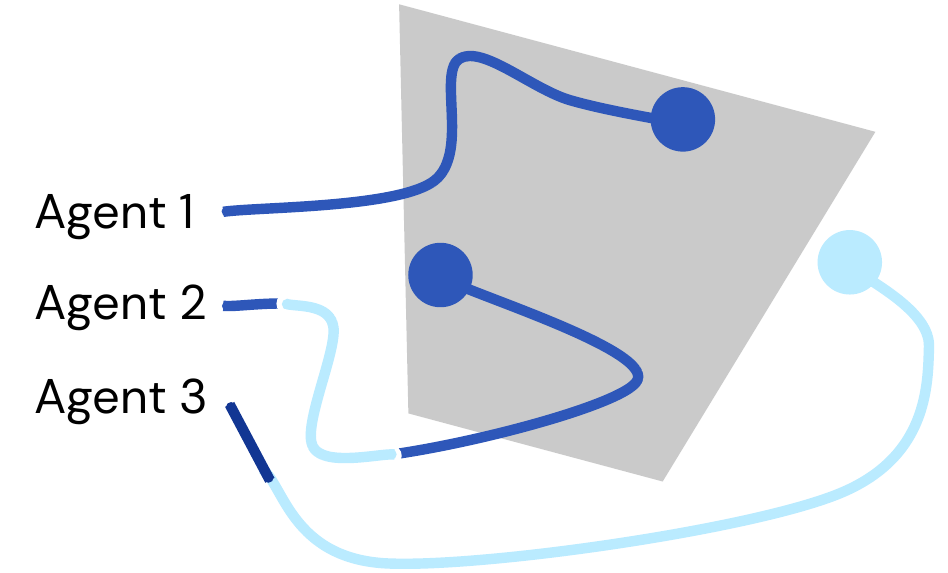}
        \caption{}
        \label{fig:imputation_overview_agents}
    \end{subfigure}%
    \hspace{30pt}
    \begin{subfigure}{0.35\textwidth}
        \centering
        \includegraphics[page=2,width=\textwidth]{figs/imputation_overview.pdf}
        \caption{}
        \label{fig:imputation_overview_masks}
    \end{subfigure}
    \caption{Stylized visualization of the multiagent time-series imputation setting. \subref{fig:imputation_overview_agents} Agent trajectories up to and including time $t$.
    Dark blue indicates trajectory portions that are observed (with light indicating otherwise); the camera field of view at the current time $t$ is indicated in grey. \subref{fig:imputation_overview_masks} Visualization of masks $\vm$ for all timesteps, where $\vm^i_t=1$ where dark, and $\vm^i_t=0$ where light. The mask at time $t$, which corresponds to the frame shown in \subref{fig:imputation_overview_agents}, is highlighted in grey.}
    \label{fig:imputation_overview}
\end{figure}

We first define the multiagent time-series imputation problem, with football as the motivating example.
As shown in \cref{fig:imputation_overview_agents}, player observations may be temporally occluded when they are out of the camera frame, and players may disappear and reappear in view multiple times throughout a sequence.
Moreover, the role of any individual player may change multiple times throughout a given trajectory sequence (e.g., a defender can behave in the manner of a midfielder or forward).
This characteristic has been well-investigated in prior works~\citep{le2017coordinated,yeh2019diverse}, and ultimately implies that learned models should be invariant to permutations of player orders within each team.
Consequently, such models should learn to predict the behavior of players conditioned on the game context, rather than purely on the players' prescribed roles in the team's formation.

More formally, consider a set of $N$ agents $\sI = \{1,\ldots,N\}$.
Let $\vx^{i}_{t} \in \R^{d}$ denote the $d$-dimensional observation of the agent $i \in \sI$ at time $t \in \sT = \{0,\ldots, T\}$.\footnote{\textbf{Notation:} We refer to any scalars associated with an agent $i$ at time $t$ as, e.g.,  $s^{i}_{t}$.
We use bold notation for vectors (e.g., $\vv^{i}_t$).
The concatenation of scalars or vectors across time and/or agent indices is denoted by, respectively, dropping the corresponding subscripts and superscripts (e.g., $\vs = s^{1:N}_{0:T}$ and $\vs_t = \vs^{1:N}_t$).
} 
In the football scenario considered in our evaluations, $d=2$, with $\vx^{i}_t$ corresponding to the $(x,y)$ position of a player or the ball on the pitch at time $t$.
For simplicity, we henceforth refer to $\vx^{i}_t$ as the state (rather than observation) of agent $i$, as it comprises the variable of interest we seek to estimate in this work.
In the time-series imputation regime, at each time step $t\in\sT$, observations may be missing for any subset of players.
Let $\vx = \vx^{1:N}_{0:T}$ be observed at the timesteps indicated by an agent-wise masking matrix $\vm$ valued in $\{0,1\}^{d}$, such that $\vm^{i}_t$ is equal to $1$ whenever  observation $i$ is available at timestep $t$, and $0$ otherwise (see \cref{fig:imputation_overview_masks}).
In the football context, each player's on-pitch $(x,y)$ position is either fully observed at a given time, or fully unobserved (i.e., no situations where their $x$-position is observed while their $y$-position is not). 
The objective is then to compute estimates $\hat{\vx} \in \R^{d}$ of all the unobserved agent states at all timesteps.
More precisely, the multiagent time-series imputation problem takes the observed states $\vx \odot \vm$ as input, where $\odot$ refers to the Hadamard product,
and aims to output a full prediction $\hat{\vx}^{1:N}_{0:T}$.
We quantify this in our experiments via the evaluation loss $\Ltwo(\hat{\vx} \odot (1-\vm), \vx \odot (1-\vm))$.

\section{Method: Graph Imputer}\label{sec:approach}

\begin{algorithm}[t]
    \caption{Graph Imputer Pseudocode}
    \label{alg:main}
    
    \begin{algorithmic}[1]
        \Function{DirectionalUpdate}{$\vx_t, \vm_t, \hat{\vx}_t$}
            \State Update autoregressively-filled state at current timestep $t$ via \cref{eq:x_autoregress}
            \State Update LSTM states via \cref{eq:bidir_lstm}
            \State Sample prior and posterior latent states via \cref{eq:gi_prior,eq:gi_encoder}
            \State $\Delta \hat{\vx}' \gets$ Sample relative state update for next timestep via \cref{eq:gi_decoder}
            \State $\hat{\vx}' \gets$ accumulate $\Delta \hat{\vx}'$ via \cref{eq:cumul_update}
            \State \Return $\hat{\vx}'$
        \EndFunction
        \Statex
        \Function{GraphImputer}{$\vx, \vm$}
            \State Initialize network parameters, initial directional estimates $\hat{\vfx}_0$ and $\hat{\vbx}_T$ to ground truth
            \For{iteration $\gets 1$ to $K$}
                \For{$t \gets 0$ to $T-1$}
                    \State $\hat{\vfx}_{t+1} \gets$ \Call{DirectionalUpdate}{$\vx_t, \vm_t, \hat{\vfx}_t$}
                \EndFor
                \For{$t \gets T$ to $1$}
                    \State $\hat{\vbx}_{t-1} \gets$ \Call{DirectionalUpdate}{$\vx_t, \vm_t, \hat{\vbx}_t$}
                \EndFor
                \State $\hat{\vx} \gets$ Fuse the forward-backward estimates $\vfx$ and $\vbx$ via \cref{eq:fusion_mean} or \cref{eq:fusion_nearest}
                \State Update model parameters via gradient step on ELBO~\cref{eq:train_elbo}
            \EndFor
            \State \Return $\hat{\vx}$
        \EndFunction
    \end{algorithmic}
\end{algorithm}

\begin{figure}[t]
    \centering
    \includegraphics[width=\textwidth,page=1]{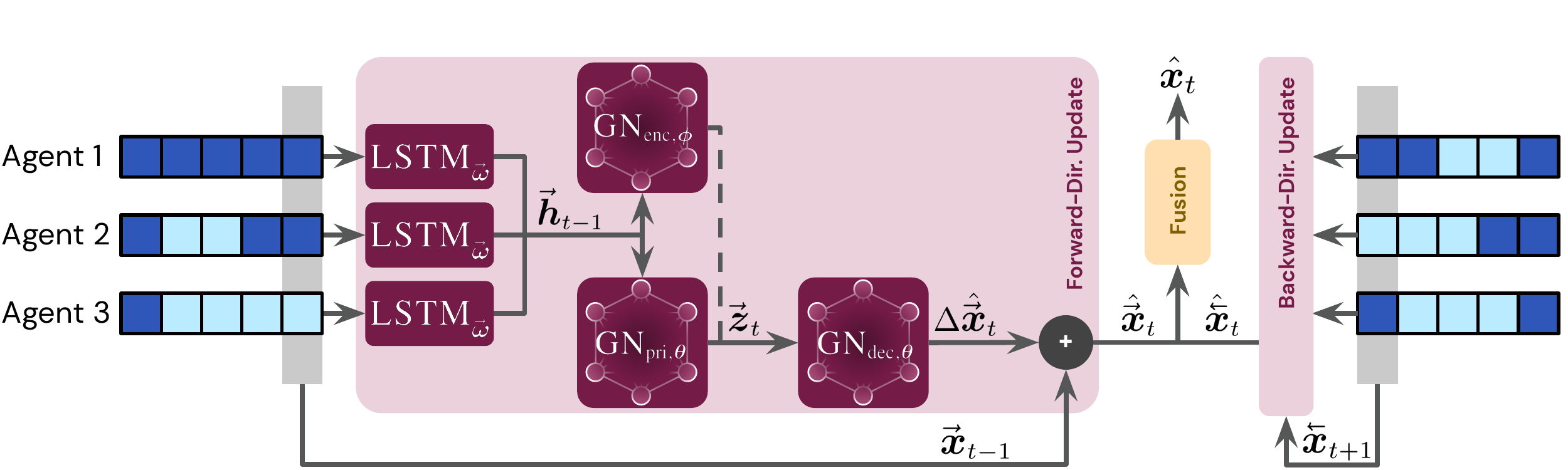}
    \caption{Graph Imputer model.
    Our model imputes missing information at each timestep using a combination of bidirectional LSTMs and graph networks.
    An exposition of a forward-direction update (corresponding to \textsc{DirectionalUpdate} in \cref{alg:main}) is provided in the left portion of the figure.
    In each direction, agent-specific temporal context is updated via LSTMs with shared parameters.
    All agents' LSTM hidden states, $\vfh_{t-1}$, are subsequently used as node features in variational graph networks to ensure information-sharing across agents.
    This enables learning of a distribution over agent state deviations, $\Delta\vfx_{t}$. 
    The process is likewise repeated in the backward-direction (right portion of the figure), with the directional updates fused to produce an imputed estimate $\hat{\vx}_t$ at each time $t$.
    }
    \label{fig:alg_overview}
\end{figure}

This section introduces our proposed approach, called the Graph Imputer.
\Cref{alg:main} provides the associated pseudocode, and \cref{fig:alg_overview} illustrates the approach at a high level.

Our approach builds on the closely related works of \citet{yeh2019diverse} and \citet{sun2018predicting}, which operate in the regime of predicting forward-rollouts of player trajectories.
The targeted individuals modeled in our domain of interest are human football players, who can exhibit stochastic behaviors on-pitch.
To enable learning of stochastic predictions given an observation stream, our underlying model effectively learns correlations along two axes: 
i) across time via bidirectional LSTMs, which autoregressively generate unobserved agent states;
and ii) across agents via a combination of graph networks~\citep{battaglia2018} and variational RNNs (VRNNs)~\citep{chung2015}, which model the multiagent interactions involved and enable sampling of distributions of imputed trajectories. 
The forward- and backward-direction imputed states are fused at each timestep, thus ensuring that all available temporal and agent-interaction information is used throughout the entire generated sequence. 
We next define the specific components of our model in detail.

\paragraph{Bidirectional autoregression.}
The key distinction between our problem regime and that of many prior multiagent predictive modeling approaches is that we target the more general imputation setting, involving both future and past contextual information about subsets of various agents.
The temporal backbone of our model is, thus, a bidirectional autoregressive LSTM, which leverages all available information at the time of prediction.

Specifically, at each time $t$, let $\vfx_t$ and $\vbx_t$ denote the forward- and backward-direction inputs to the model.
These inputs correspond to the combination of ground truth states, $\vx_t$, for observed agents, and autoregressively-predicted states, $\hat{\vfx}_t$ and $\hat{\vbx}_t$ (defined below), for unobserved agents, as follows: 
\begin{align}\label{eq:x_autoregress}
    \vfx_t &= \vx_t \odot \vm_t + \hat{\vfx}_t \odot (1-\vm_t) \quad &
    \vbx_t &= \vx_t \odot \vm_t + \hat{\vbx}_t \odot (1-\vm_t) \, .
\end{align}
We use bidirectional LSTMs to temporally-integrate observation sequences and learn the forward- and backward-dynamics involved.
Agent-wise hidden states, $\vfh_t^i$ and $\vbh_t^i$, are updated as follows:
\begin{align}\label{eq:bidir_lstm}
    \phantom{\quad \forall i \in \sI}
    \vfh_t^i = \LSTM_{\vfomega}(\vfx_t^i, \vfh^i_{t-1})
    \qquad\qquad\qquad
    \vbh_t^i = \LSTM_{\vbomega}(\vbx_t^i, \vbh^i_{t+1})\, \quad \forall i \in \sI \,,
\end{align}
where $\vfomega$ and $\vbomega$ refer to direction-specific LSTM parameters, which are shared across agents. 

We next detail the computation of the autoregressively-predicted states $\hat{\vfx}$ and $\hat{\vbx}$ appearing in \cref{eq:x_autoregress}, which are sampled from a variational graph network capturing multiagent interactions in the system. 

\paragraph{Graph Networks.} 
We define a graph network consisting of $N$ nodes, each corresponding to an agent or entity in the system (e.g., $N=23$ in the football domain, capturing the state of the 22 players and the ball).
Let $\vv^i_t$ denote the node feature vector associated with an agent $i \in \sI$, which encodes its spatiotemporal context at time $t$. 
Likewise, let $\ve^{(i,j)}$ denote the directed edge feature connecting agent $i \in \sI$ to agent $j \in \sI$.
Graph networks operate via rounds of message passing, which update edge and node features to propagate information across the various nodes involved.
In our instance, the message-passing update is expressed as follows,
\begin{align}
    \ve'^{(i,j)} & = f_{\vtheta}^e(\vv^i, \vv^j)  \quad & \text{(Update edges from sender nodes $i \in N^{-}(j)$ to recipients $j \in \sI$)}\label{eq:gn_edge_up}\\
    \ve'^j &= \sum_{i \in N^{-}(j)} \ve'^{(i,j)} \quad &\text{(Aggregate incoming edges for all receiver nodes $j \in \sI$)}\label{eq:gn_edge_agg}\\
    \vv'^{j} & = f_{\vtheta}^v(\ve'^j) \quad &\text{(Update all receiver nodes $j \in \sI$)}\label{eq:gn_node_up}\, ,
\end{align}
where $N^{-}(j)$ are in-neighbors of node $j$, and $f_{\vtheta}^e$ and $f_{\vtheta}^v$ are, respectively, edge and node update functions with learned parameters $\vtheta$.
In shorthand, given an initial set of node features $\vv$, we refer to the updated features following the message-passing steps in \cref{eq:gn_edge_up,eq:gn_edge_agg,eq:gn_node_up} as $\vv' = \GN_{\vtheta}(\vv)$.

\paragraph{Variational updates.}
At any time $t$, the history of autoregressively-filled directional inputs, $\vfx_{<t}$ and $\vbx_{>t}$, is encoded by the LSTM states $\vfh_{t-1}$ and $\vbh_{t+1}$.
Conditioned on this context, our model uses variational graph networks to enable information-sharing across agents, and learn a distribution over latent random variables $\vz$ and predicted state updates $\Delta\hat{\vx}_t$.
Specifically, the graph imputer learns to approximate the directional prior distributions $p_{\vtheta}(\vfz^i_t|\cdot)$ and $p_{\vtheta}(\vbz^i_t|\cdot)$, posterior distributions $q_{\vphi}(\vfz^i_t|\cdot)$ and $q_{\vphi}(\vbz^i_t|\cdot)$, and decoded output distribution $p_{\vtheta}(\Delta\hat{\vfx}^i_t|\cdot)$ and $p_{\vtheta}(\Delta\hat{\vbx}^i_t|\cdot)$, as follows,
{
\mathtight
\begin{align}
    p_{\vtheta}(\vfz_t^i | \vfx_{<t}, \vfz_{<t}) &=  \mathcal{N}\left( \vfmu_{\pri,t}^i, (\vfsigma_{\pri,t}^{i})^2 \right)  
    \hspace{4pt}&
    p_{\vtheta}(\vbz_t^i | \vbx_{>t}, \vbz_{>t}) &=  \mathcal{N}\left( \vbmu_{\pri,t}^i, (\vbsigma_{\pri,t}^{i})^2 \right)\label{eq:gi_prior}\\
    q_{\vphi}(\vfz_t^i | \vx_t, \vfx_{< t}, \vfz_{<t}) &=  \mathcal{N}\left( \vfmu_{\enc,t}^i, (\vfsigma_{\enc,t}^{i})^2 \right)  
    &
    q_{\vphi}(\vbz_t^i | \vx_t, \vbx_{> t}, \vbz_{>t}) &=  \mathcal{N}\left( \vbmu_{\enc,t}^i, (\vbsigma_{\enc,t}^{i})^2 \right)\label{eq:gi_encoder}\\
    p_{\vtheta}(\Delta \hat{\vfx}_t^i | \vfx_{<t}, \vfz_{\leq t}) &=  \mathcal{N}\left( \vfmu_{\dec,t}^i, (\vfsigma_{\dec,t}^{i})^2 \right)  
    &
    p_{\vtheta}(\Delta \hat{\vbx}_t^i | \vbx_{>t}, \vbz_{\geq t}) &=  \mathcal{N}\left( \vbmu_{\dec,t}^i, (\vbsigma_{\dec,t}^{i})^2 \right)\, .\label{eq:gi_decoder}
\end{align}
}%
In the above, \cref{eq:gi_prior} enables sampling of latent variables, $\vfz_t$ and $\vbz_t$, conditioned on the prior  information available up to, though not including, the prediction timestep $t$. 
Likewise, \cref{eq:gi_encoder} captures the posterior latent state distribution, conditioned on the same information as the prior \emph{in addition to} the ground truth state $\vx_t$.
Finally, \cref{eq:gi_decoder} enables sampling of a next-state prediction for each direction.
As in typical VRNN-based approaches, the encoder is used only during training to sample latent states $\vz_t$, which are used as inputs for the decoder;
during evaluation, samples $\vz_t$ from the prior are used instead, as the encoder can, naturally, no longer be used due to the ground truth state $\vx_t$ being unavailable.

The collection of mean and variance parameters above, $\vmu$ and $\vsigma^2$, parameterize underlying Gaussian distributions. 
These parameters simply correspond to node features output by underlying graph networks, which exchange information between agents following a message-passing step: 
{
\mathtight
\begin{align}
    \left[\vfmu_{\pri,t}^i, (\vfsigma_{\pri,t}^{i})^2\right]_{i \in \sI} &= \GN_{\pri,\vtheta}\!\!\!\left(\vfh_{t-1}\right)
    \hspace{25pt} &
    \left[\vbmu_{\pri,t}^i, (\vbsigma_{\pri,t}^{i})^2\right]_{i \in \sI} &= \GN_{\pri,\vtheta}\!\!\!\left(\vbh_{t+1}\right)\label{eq:gn_prior}\\
    \left[\vfmu_{\enc,t}^i, (\vfsigma_{\enc,t}^{i})^2\right]_{i \in \sI} &= \GN_{\enc,\vphi}\!\!\!\left(\left[\vx_t, \vfh_{t-1} \right]\right)
    &
    \left[\vbmu_{\enc,t}^i, (\vbsigma_{\enc,t}^{i})^2\right]_{i \in \sI} &= \GN_{\enc,\vphi}\!\!\!\left(\left[\vx_t, \vbh_{t+1} \right]\right)\label{eq:gn_encoder}\\
    \left[\vfmu_{\dec,t}^i, (\vfsigma_{\dec,t}^{i})^2\right]_{i \in \sI} &= \GN_{\dec,\vtheta}\!\!\!\left(\left[\vfz_t, \vfh_{t-1} \right]\right)
    &
    \left[\vbmu_{\dec,t}^i, (\vbsigma_{\dec,t}^{i})^2\right]_{i \in \sI} &= \GN_{\dec,\vtheta}\!\!\!\left(\left[\vbz_t, \vbh_{t+1} \right]\right)\,. \label{eq:gn_decoder}
\end{align}
}%
Subsequent to their sampling in \cref{eq:gi_decoder}, the relative (delta) state updates, $\Delta \hat{\vx}_t$, are accumulated to produce predictions in absolute-space, 
\begin{align}\label{eq:cumul_update}
    \hat{\vfx}_t&= \vfx_{t-1} + \Delta\hat{\vfx}_t
    &
    \hat{\vbx}_t&= \vbx_{t+1} + \Delta\hat{\vbx}_t\, .
\end{align}
These predicted states $\hat{\vfx}_t$ and $\hat{\vbx}_t$ are then used to autoregressively update the next-timestep inputs using $\cref{eq:x_autoregress}$.
The procedure then continues to autoregressively update the states for all timesteps $t$ in each respective direction.

\paragraph{Forward-backward fusion.} 
The final directional outputs from the model are subsequently fused to produce the bidirectional estimates $\hat{\vx}^i_t$ for all agents.
As in recent works on bidirectional LSTM-based imputation~\citep{cao2018_brits}, one method of fusion is to simply take the mean, 
\begin{align}\label{eq:fusion_mean}
    \hat{\vx}^i_t = 0.5(\hat{\vfx}^i_t + \hat{\vbx}^i_t) \, .
\end{align}
Alternatively, at time $t$, let $\ftau^i_t$ and $\btau^i_t$ denote the number of timesteps until the next ground truth observation in each direction, respectively.
One can then weigh the contribution of each direction as,
\begin{align}\label{eq:fusion_nearest}
    \hat{\vx}^i_t = (\ftau^i_t \hat{\vfx}^i_t + \btau^i_t \hat{\vbx}^i_t)(\ftau^i_t + \btau^i_t)^{-1} \, .
\end{align}
This ensures predictions corresponding to the direction with the most recent observation are weighted higher.
For example, if at prediction timestep $t$, the nearest ground truth observations in the future and past for agent $i$ occur at $t+8$ and $t-2$, then $\ftau^i_t = 8$ and $\btau^i_t = 2$, such that $\hat{\vx}^i_t = 0.8\hat{\vfx}^i_t + 0.2\hat{\vbx}^i_t$.

\paragraph{Training.} As in prototypical VAE pipelines, we update model parameters in each iteration of the algorithm by maximizing the evidence lower bound (ELBO) over all the agents in each trajectory,
{
\mathtight
\begin{align}\label{eq:train_elbo}
    \sum_{t\in\sT} 
    \bigg[
    &\E_{
    q_{\vphi}(\vfz_t | \vx_t, \vfx_{< t}, \vfz_{<t}) 
    }\left[ 
    \log p_{\vtheta}(\Delta \hat{\vfx}_t | \vfx_{<t}, \vfz_{\leq t})\right]
    - \beta D_{KL}(
        q_{\vphi}(\vfz_t | \vx_t, \vfx_{< t}, \vfz_{<t}) 
        ||
        p_{\vtheta}(\vfz_t | \vfx_{<t}, \vfz_{<t}) 
    )
     + \nonumber\\
    &\E_{
    q_{\vphi}(\vbz_t | \vx_t, \vbx_{> t}, \vbz_{>t})
    }\left[ 
        \log 
        p_{\vtheta}(\Delta \hat{\vbx}_t | \vbx_{>t}, \vbz_{\geq t}) \right]
        - \beta D_{KL}(
            q_{\vphi}(\vbz_t | \vx_t, \vbx_{> t}, \vbz_{>t})
            ||
            p_{\vtheta}(\vbz_t | \vbx_{>t}, \vbz_{>t}) 
        )
    \bigg] \, ,
\end{align}
}%
where $\beta$ is a weighing term on the VAE KL-regularizer~\citep{higgins2017beta}.
For training, we maximize \cref{eq:train_elbo} over mini-batches of trajectories sampled from our dataset.

In our experiments, we also consider several ablations of the models, including:
decoders that take as input only the latent states $\vfz_t$ and $\vbz_t$ (i.e., disabling the \textbf{skip-connection} from the LSTM hidden states $\vfh_{t-1}$ and $\vbh_{t+1}$ to the decoder in \cref{eq:gn_decoder});
and \textbf{next-step conditioned} graph-decoders that include nodes with features $\vv^{i}$ locked to agent observations available for the timestep being predicted  (i.e., observed decoder nodes with features $\vx^i_t \odot \vm^i_t$, which only send messages during message-passing, and thus do not update their states at prediction timestep $t$ as they are observable).

\section{Evaluation}\label{sec:evaluation}
We next empirically evaluate the Graph Imputer against a range of existing models.

\paragraph{Dataset.}
We use a dataset of 105 English Premier League matches, where all on-pitch players and the ball are tracked at 25~frames-per-second for each match.
We partition the data into trajectory sequences of 240 frames (each capturing 9.6$s$ of gameplay), then downsample the data to 6.25~frames-per-second.
For training purposes, we retain only trajectories with 22 players available in the raw data (such that we can compute losses against all players' ground truth),
and spatially realign the data such that the team in possession always moves towards the right of the pitch (as done in prior works~\citep{bialkowski2014large}). 
Finally, for training and evaluation, we split the resulting data into two partitions of 30838 and 4024 trajectories, respectively.
While we are not permitted to release this data due to the licensing terms involved, tracking data of similar specifications is available from existing public sources.\footnote{For example, see \url{https://github.com/metrica-sports/sample-data} and \url{https://github.com/Friends-of-Tracking-Data-FoTD/Last-Row} for tracking data released by other authors.}

\paragraph{Simulated camera model.}
We use a simulated camera model to generate a realistically-structured mask for the task of off-screen player trajectory imputation. 
The camera model is parameterized by its position and horizontal and vertical field of view angles, with the parameters chosen to produce a vantage point similar to a stadium broadcast camera.
For simplicity, the camera-normal is set to track the ball position at each timestep.
By intersecting the camera view cone with the pitch plane, we obtain the projected in-frame polygon and mask out-of-frame players accordingly (as in \cref{fig:imputation_overview}).
On average, $12.76 \pm 3.70$ players (out of 22) are in-frame in each sequence, with a consecutive in-frame duration of $4.94s \pm 3.49s$.
Under this camera model, certain players are at times completely out of view for the entire trajectory duration.
To provide some warm-up context to the models during training, we include an additional $5$ frames of observations at the beginning and end of all trajectories for all players.
In practical evaluation settings involving longer trajectory sequences, the camera pans around such that all players are effectively observed at some stage, thus not requiring this.

\paragraph{Baselines.}
We compare our approach against the following baselines. 
\textbf{Linear}: A baseline consisting of linearly interpolating players' positions from the moment they leave the the camera field of view to the moment they return, thus adhering to boundary value constraints.
\textbf{Autoregressive LSTMs}: A simple baseline using autoregressive LSTMs, run independently per player for state estimation.
\textbf{Role-invariant VRNNs}: A strong variational baseline hand-crafted for the football scenario (i.e., assuming two teams of an equal number of players), using VRNNs and a combination of post-processing steps to ensure information-sharing between players on each team, and invariance of model outputs to re-ordering of players in inputs. Refer to \cref{sec:appendix_exp_setup} for details. 
\textbf{Social LSTM}~\citep{Alahi_2016_CVPR}: A model that uses social pooling to ensure spatially-nearby context is appropriately shared between individual agents. We also train a bidirectional Social LSTM variant using a combination of the vanilla Social LSTM updates and the fusion equations \cref{eq:x_autoregress,eq:bidir_lstm,eq:fusion_mean,eq:fusion_nearest}, which we have not observed being used in the literature for our problem regime.
\textbf{GVRNNs}~\citep{yeh2019diverse}: A model that uses a combination of VRNNs and Graph Neural Networks (similar in nature to Graph-VRNNs~\citep{sun2018predicting}).

\paragraph{Training and Hyperparameters.}
We conduct a wide hyperparameter sweep and report the results corresponding to the best hyperparameters for each model.
We train for $1e5$ iterations, with a  batch size of $64$ trajectories, using the Adam 
optimizer~\citep{kingma2014adam} with a learning rate of $1e-3$ (and default exponential decay parameters, $b_1=0.9$, and $b_2=0.999$).
For LSTM-based models (including the ones used in the Graph Imputer), we use 2-layer LSTMs with 64 hidden units each.
For the graph edge and node update networks, $f_{\vtheta}^e$ and $f_{\vtheta}^v$, we use 2-layer MLPs with 64 hidden units each, with internal ReLU activations~\citep{NairH10}.
In the ELBO~\cref{eq:train_elbo}, we anneal $\beta$ from an initial value of $0.1$ to final values of $0.01$ and $1$ in our sweeps.
For all bidirectional models, we sweep over the two fusion modes specified in \cref{eq:fusion_mean,eq:fusion_nearest}. For each model, training for each hyperparameter set is conducted and reported over a sweep of 5 random seeds.
Additional hyperparameters and computational resources used are detailed in \cref{sec:appendix_exp_setup}.

\paragraph{Results.}

\begin{table}[t]
\centering
\setlength\tabcolsep{4.8pt} 
\caption{Football off-screen player state estimation results.
The columns refer to the following:
\textbf{Football-specific}: whether the model is hand-crafted for the football case and not directly applicable to general multiagent domains (i.e., processes data in a manner explicitly assuming two teams of players, along with a ball).
\textbf{Skip connection}: whether a skip-connection from the input to the decoder is enabled for autoencoder based models. 
\textbf{Next-step cond. decoder}: whether decoders in graph network-based models condition on available next-timestep observations, as additional context.
For each baseline model, we compute the mean evaluation loss, $\Ltwo(\text{Mean})$, compared to the ground truth trajectories (over all seeds).
For stochastic models, for each evaluation sequence we also take $6$ samples of imputed trajectories, and also report the minimum evaluation loss, $\Ltwo(\text{Min.})$, over all samples, averaged over all seeds.
}
\begin{tabular}{P{3.95cm}
P{1.1cm}
P{1.6cm}
P{1.4cm}
S[table-format=1.3]@{\,\( \pm \)\,}
S[table-format=1.3]
S[table-format=1.3]@{\,\( \pm \)\,}
S[table-format=1.3]}
\toprule 
Model & Football-specific & Skip connection & Next-step cond. decoder & \multicolumn{2}{c}{$\Ls_{2}(\text{Mean})$} & \multicolumn{2}{c}{$\Ls_{2}(\text{Min.})$}\\
 \midrule
                 Role-invariant VRNN &   \cmark &  \xmark &     $-$ &           2.020 &            2.03 &           1.960 &           2.063 \\
                 Role-invariant VRNN &   \cmark &   \cmark &     $-$ &           0.958 &           0.009 &           0.953 &           0.009 \\
          Bidir. Role-invariant VRNN &   \cmark &  \xmark &     $-$ &           0.174 &           0.002 &           0.160 &           0.002 \\
          Bidir. Role-invariant VRNN &   \cmark &   \cmark &     $-$ &           0.167 &           0.002 &           0.166 &           0.002 \\
                              \midrule Linear &  \xmark &     $-$ &     $-$ &           0.658 &           0.081 &             \multicolumn{2}{c}{$-$} \\
                                LSTM &  \xmark &     $-$ &     $-$ &           1.579 &           0.019 &             \multicolumn{2}{c}{$-$} \\
                         Bidir. LSTM &  \xmark &     $-$ &     $-$ &           0.350 &           0.006 &             \multicolumn{2}{c}{$-$} \\
 Social LSTM~\citep{Alahi_2016_CVPR} &  \xmark &     $-$ &     $-$ &           1.049 &           0.274 &             \multicolumn{2}{c}{$-$} \\
                  Bidir. Social LSTM &  \xmark &     $-$ &     $-$ &           0.198 &           0.052 &             \multicolumn{2}{c}{$-$} \\
        GVRNN~\citep{yeh2019diverse} &  \xmark &  \xmark &  \xmark &           2.243 &           0.136 &           1.453 &           0.073 \\
        GVRNN~\citep{yeh2019diverse} &  \xmark &  \xmark &   \cmark &           2.447 &           1.197 &           2.400 &           1.231 \\
        GVRNN~\citep{yeh2019diverse} &  \xmark &   \cmark &  \xmark &           0.882 &           0.009 &           0.874 &           0.009 \\
        GVRNN~\citep{yeh2019diverse} &  \xmark &   \cmark &   \cmark &           0.865 &           0.018 &           0.852 &           0.017 \\
                Graph Imputer (Ours) &  \xmark &  \xmark &  \xmark &           0.241 &            0.05 &           0.224 &           0.051 \\
                Graph Imputer (Ours) &  \xmark &  \xmark &   \cmark &           0.404 &           0.102 &           0.397 &            0.11 \\
                Graph Imputer (Ours) &  \xmark &   \cmark &  \xmark &           0.165 &           0.005 &           0.163 &           0.005 \\
                Graph Imputer (Ours) &  \xmark &   \cmark &   \cmark &  \textbf{0.153} &  \textbf{0.003} &  \textbf{0.151} &  \textbf{0.003} \\
\bottomrule
\end{tabular}
\label{table:results}
\end{table}

\begin{figure}[t]
    \setcounter{row}{1}%
    \setcounter{subfigure}{0}%
    \centering
    \begin{tabularx}{\textwidth}{p{0.0in}YYY}
        & \hspace{15pt} Example 1 & \hspace{15pt} Example 2 & \hspace{15pt} Example 3\\
        \resultsRow{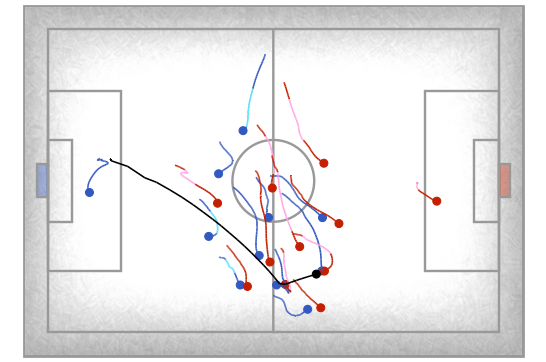}{Ground Truth Only}{0}
        \resultsRow{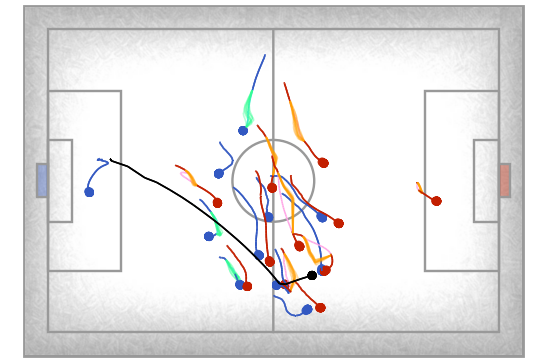}{GVRNN}{0}
        \resultsRow{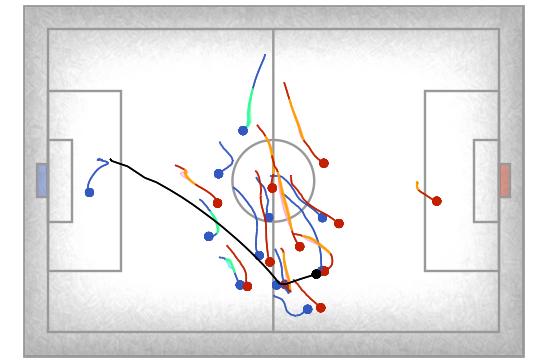}{Graph Imputer}{0}
    \end{tabularx}
    \hspace*{25pt}\includegraphics[trim={0 0 0 25pt}, width=0.95\textwidth]{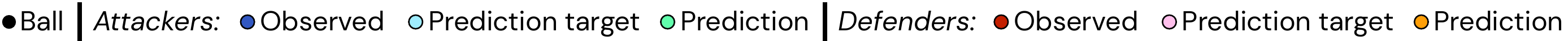}
    \caption{Trajectory visualizations (best viewed when zoomed in).
    Each column provides an example trajectory sequence, with the first row illustrating the ground truth, and subsequent rows showing results from various models, including the Graph Imputer (ours).
    For all examples, the Graph Imputer trajectories seamlessly adhere to the boundary value constraints imposed at the moments of disappearance and reappearance of players.
    }
    \label{fig:traj_examples}
\end{figure}

\Cref{table:results} provides a summary of results for the football off-screen player data imputation regime, including ablations over key model features where applicable.
As noted earlier, the role-invariant models (listed in the first several table rows) are hand-crafted for the football case, and thus are not applicable to general multiagent settings; 
nonetheless, these models pose a strong evaluation baseline, and outperform several of the more generic approaches. 
Our proposed model, the Graph Imputer, outperforms the baselines both in terms of the mean and minimum evaluation loss over prediction samples, including the hand-crafted models. 

As evident in \cref{table:results}, bidirectionality naturally yields a significant improvement in terms of overall performance across the models, as both past and future information is used in estimating player positions when off-screen.
This is quantitatively evident even for the linear baseline, which is effectively bidirectional as it interpolates the last appearance and first reappearance of each player.
The unidirectional Social LSTM model is outperformed by the strongest-performing GVRNN model (which uses both a skip connection and next-step conditioned graph decoder), as observed in earlier literature. 
However, as the Social LSTM is fundamentally deterministic in nature, it cannot be used for sampling multiple viable player trajectories.
We additionally observe that use of a skip connection from inputs to the decoder results in a significant improvement in results, for all variational models considered.
While use of a next-step conditioned graph decoder slightly improves results for the Graph Imputer, it has a more significant impact on the GVRNN model, which we conjecture is due to the former model's bidirectional nature already providing significant information about future observations.

\Cref{fig:traj_examples} provides static visualizations of trajectory results for several example sequences, with additional examples in \cref{sec:appendix_results}. 
\ifarXiv
    We recommend readers view our animated visualizations on our website.\footnote{\url{https://sites.google.com/view/imputation-of-football/}}
\fi
These animations illustrate the simulated camera model and provide the most intuitive means of visualizing the results due to their spatiotemporal nature.

In \cref{fig:traj_examples}, observed trajectory segments for the attacking and defending team are, respectively, illustrated in dark blue and red, with the ball trajectory indicated in black.
In the first row of the figure, we illustrate the portion of player trajectories that are unobserved in light blue and pink for each team, respectively.
Recall that the observations provided to the models are the raw positions available for in-camera players, with the camera tracking the ball in each timestep.
Well-performing models will, ideally, learn the key behavioral characteristics of player interactions and physics (e.g., velocities, constraints on acceleration, player turning radii, etc.) given the available positional information to make realistic predictions.
The subsequent rows illustrate the predictions made by both the GVRNN model and the Graph Imputer, under the same observations.
Notably, the bidirectional nature of our Graph Imputer approach enables predictions to not only more accurately model the flow of movement of players on the pitch, but also to appropriately adhere to the boundary value constraints imposed by players when they appear back in the camera view. 
For additional experiment results, including numerous visualizations over more baseline models and ablations over the bidirectional fusion modes \cref{eq:fusion_mean,eq:fusion_nearest}, refer to \cref{sec:appendix_results}.

\section{Related Work}\label{sec:related_work}
There exist a number of works from various fields of research that are related to our approach. 
Specifically, a number of works from robotics and computer vision~\citep{Alahi_2016_CVPR,Sakata18,Rudenko20,zhu2021learning}, sports analytics~\citep{Suda19,le2017data,yeh2019diverse,hauri2020multimodal,alcorn2021baller2vec}, economics~\citep{Taylor07,Sezer20,tay2001application} and machine learning~\citep{FoersterFANW18,BrownLGS19,sun2018predicting} focus on various combinations of missing data imputation and multiagent trajectory predictions. 
Given the broad scope of time series prediction as a research field~\citep{han2019review}, we focus particularly on models that predict human trajectories~\citep{Rudenko20}, as they are the most relevant for our problem regime. 
We also provide an in-depth cross-sectional table of the most-closely related models in \cref{sec:appendix_related_works}.  

One of the most common applications within human trajectory prediction (albeit not directly related to sports), is pedestrian modeling~\citep{Alahi_2016_CVPR, Sun_2020_CVPR}.  
More closely related to our work are models that predict the trajectories of athletes in a team, such as basketball~\citep{li2020Evolvegraph, SuHSP17, alcorn2021baller2vec,alcorn2021b} or football~\citep{le2017data, le2017coordinated, yeh2019diverse}. 
Efforts that focus on the latter vary in the way that they treat the interactions between players. While some directly use the information about the players as conditioning for imitation learning~\citep{le2017data, le2017coordinated} others use more complex interaction models such as graph networks for forward-prediction~\citep{yeh2019diverse,sun2018predicting}.
Finally, despite being framed as a supervised learning problem rather than sequence prediction, \citet{hoshen2017} also take into account the interactions between the different variables in their multivariate trajectory prediction problem by using interaction networks to model them. 

Rather than targeting the forward-prediction regime, the  goal of our model is to carry out imputation of incomplete time series involving multiple interacting agents. 
Imputation of sequential data itself can be treated as a means to an end for a separate task such as classification~\citep{che2018recurrent}. 
Also related to our line of work is prior work on a bidirectional model that carries out trajectory imputation~\citep{liu2019naomi}. 
Unlike ours, however, their approach does not target specifically the multiagent setting, though applies to the regime by essentially treating it as a large single-agent scenario.  
Finally, approaches that focus more directly on the imputation task itself include GAN-based models~\citep{luo2018multivariate,luo2019e2gan} and bidirectional inference models~\citep{cao2018_brits, DBLP:journals/corr/abs-1711-08742}. 

\section{Discussion}\label{sec:discussion}
We introduced a technique for multiagent time-series imputation, called the Graph Imputer.
Our approach uses a combination of bidirectional recurrent models to ensure use of all available temporal information, and graph networks to model inter-agent relations. 
We illustrated that our approach outperforms several state-of-the-art methods on a large dataset of football tracking data, and qualitatively yields trajectory samples that capture player interactions and adhere to the constraints imposed by available observations.
The key benefit of our approach is its generality, in the sense that it permits any subset of agents to be unobserved at any timestep, works with temporal occlusions of arbitrary time horizons, and can apply directly to general multiagent domains beyond football.

\paragraph{Limitations.} Nonetheless, there are several associated limitations of note.
Namely, the forward and backward-direction latent vectors, $\vfz$ and $\vbz$, are sampled independently in our model; 
sampling these from a joint underlying distribution could significantly improve correlations in the directional predictions.
Moreover, our model requires observations of each agent for at least a single timestep throughout each trajectory. While this is not a major limitation given long enough trajectory sequences in practice, investigating a means of enabling the model to seamlessly handle \emph{completely missing} agents would increase its generalizability.

\paragraph{Societal impacts.}
The increasing performance of predictive models such as ours also necessitates consideration of potentially negative societal impacts.
Specifically, while our approach is motivated by the use-case of sports trajectory prediction, it could also be applied to, e.g., modeling of pedestrian trajectories in the real world. 
In combination with facial recognition technologies, tracking approaches could foreseeably be used by adversaries to make predictions about and impede movements of unaware individuals.
Nonetheless, for the time being, ours and related approaches are inherently limited to reasonably short prediction time horizons and work best in well-controlled environments, and could be limited by the high stochasticity of human movement in less controlled environments. 
This makes the application of our approach for such adversarial scenarios more difficult, though certainly worth investigating and mitigating in the future as the performance of such models increases. 


\ifarXiv
    \section*{Acknowledgements}
    The authors thank Mark Rowland and Remi Munos for their helpful feedback on the paper draft.
    The authors are also grateful to the DeepMind football analytics team for their help and support on the overall research effort, including particularly Alexandre Galashov, Nathalie Beauguerlange, Jackson Broshear, and Demis Hassabis.
\else
\fi

\bibliographystyle{plainnat}
\bibliography{references}

\clearpage
\ifarXiv
\fi




\newpage

\renewcommand{\thepage}{}

\appendixpageoff
\appendixtitleoff
\renewcommand{\appendixtocname}{Supplementary material}

\ifarXiv \else \begin{bibunit} \fi

\begin{appendices}
\crefalias{section}{supp}


\renewcommand{\thefigure}{S\arabic{figure}}    
\renewcommand{\thetable}{S\arabic{table}}  
\setcounter{figure}{0}    
\setcounter{table}{0}

\begin{center}
    \textbf{\Large{Supplementary Material: \papertitle}}
\end{center}

We provide here supplementary material that may be of interest to the reader.
Note that sections and figures in the main text that are referenced here are clearly indicated via numerical counters (e.g., \cref{fig:imputation_overview}), whereas those in the appendix itself are indicated by alphabetical counters (e.g., \cref{fig:role_invariant_vrnn}).

\section{Additional Experiment Details}\label{sec:appendix_exp_setup}

\subsection{Baseline Model Details}\label{sec:appendix_baselines}
This section details the Role-invariant VRNN baselines presented in \cref{table:results} of the main paper.
As mentioned earlier, this model is designed specifically for the case of football, assuming that each trajectory stream consists of two teams (where permutation-invariance to player ordering is desired within each) and the ball. 

\begin{figure}[h]
    \centering
    \includegraphics[width=\textwidth]{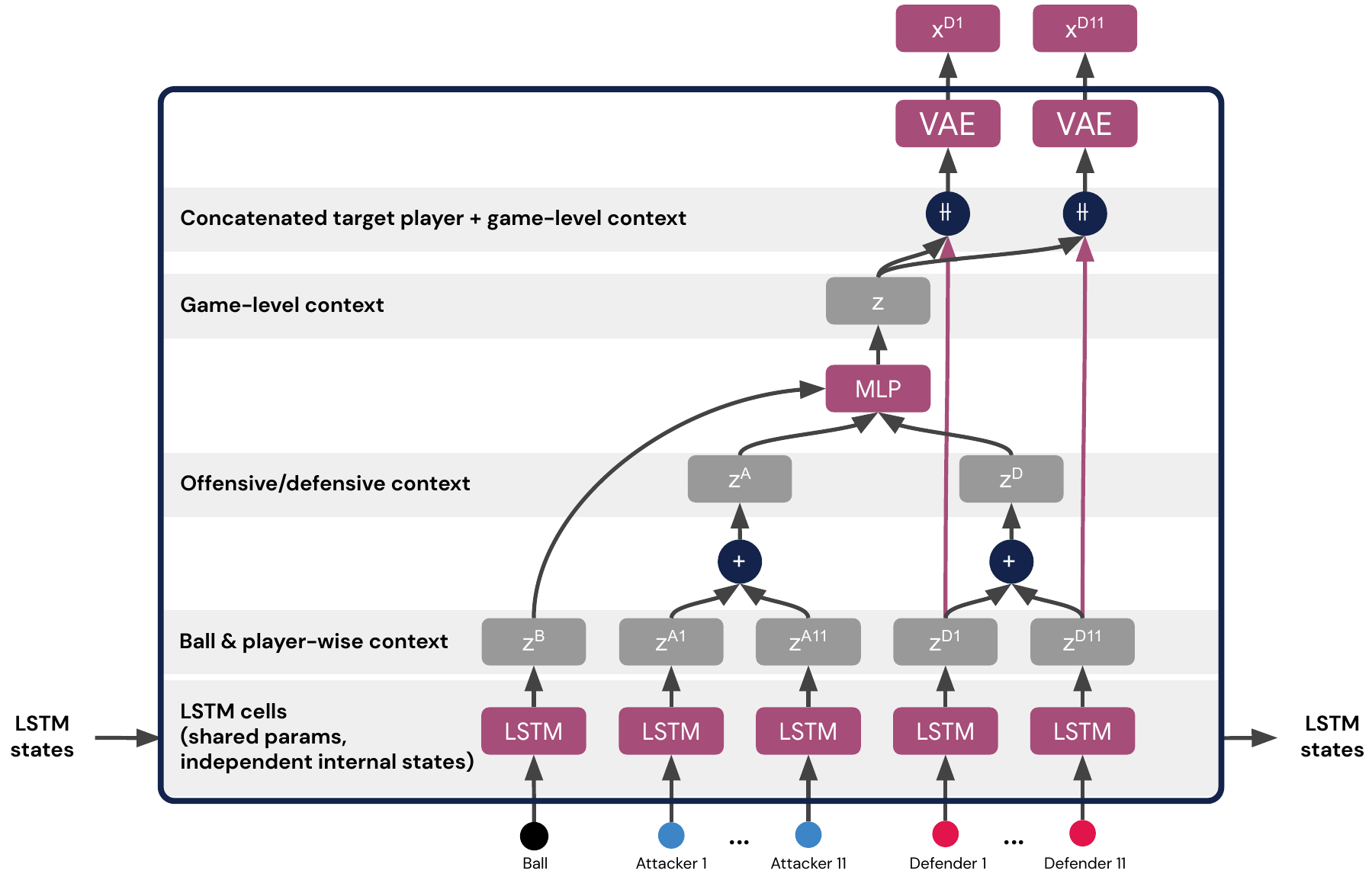}
    \caption{Football-specific role-invariant VRNN baseline architecture. The input manipulations conducted in this model, which are detailed in \cref{sec:appendix_baselines}, ensure invariance of outputs to permutations of player orders within each time. However, this also implies the strongest assumption on the domain at hand, which is interaction of two teams of players, along with a singleton entity (the ball), in a shared environment.}
    \label{fig:role_invariant_vrnn}
\end{figure}

We provide an overview of this model in \cref{fig:role_invariant_vrnn}.
At each timestep, this model works by passing the player and ball observations through LSTMs with shared parameters.
The hidden states from the LSTMs are subsequently summed up within each team, thus producing team-level contextual vectors; note that this operation ensures permutation-invariance within each team.
Subsequently, the ball and each team's hidden state are passed through an MLP, thus producing a vector representing the overall game-level context.
Finally, to produce predictions for individual players, this game-level context is concatenated to their individual LSTM states to produce a player-specific context, which is then passed through an MLP-based VAE, enabling sampling of players' next-states.

The process iterates autoregressively as in the Graph Imputer, and likewise can be repeated in reverse temporal fashion to produce and fuse bidirectional player state estimates. 
Training of this model is conducted in the same manner as the Graph Imputer, using the ELBO \cref{eq:train_elbo}.

In addition to this model, we also include in \cref{table:appendix_results} of the supplementary materials a variant called `Role-invariant RNN', which simply replaces the VAE head in \cref{fig:role_invariant_vrnn} with an MLP.

\subsection{Additional Hyperparameter Details \& Computational Resources}

In addition to the key hyperparameters detailed in the main paper, we also ran sweeps for VAE-based models wherein a standard normal prior distribution was used (in lieu of a learned prior), as typically also considered in VAE approaches.
For the Social LSTM model, we also ran sweeps over grid widths $8$, $24$, and $64$ capturing the size of neighbor grids for each player (in meters); larger grid sizes correspond to increasing amounts of neighbor context on the football pitch.

For training, we use a cluster of Tesla V100 and P100 GPUs for training and evaluation, respectively.
Overall, our sweeps were conducted over a set of 435 independent training runs (i.e., each with a unique hyperparameter set and random seed).
Depending on the simplicity of the underlying model (simplest being the autoregressive LSTM, and most complex being the Graph Imputer), each training run took approximately 3 to 15 hours of wallclock time to train.

\section{Additional Experiment Results}\label{sec:appendix_results}

\afterpage{%
\clearpage
\thispagestyle{empty}
\begin{landscape}
\begin{table}[t]
\centering
\setlength\tabcolsep{4.8pt} 
\caption{Football off-screen player state estimation results.
The columns refer to the following:
\textbf{Football-specific}: whether the model is hand-crafted for the football case and not applicable to general multiagent domains (i.e., processes data in a manner explicitly assuming two teams of players, along with a ball).
\textbf{Skip connection}: whether a skip-connection from the input to the decoder is enabled for autoencoder based models. 
\textbf{Next-step cond. decoder}: whether decoders in graph network-based models condition on available next-timestep observations, as additional context.
\textbf{Bidir. fusion mode}: the fusion mode used for bidirectional models, where `mean' corresponds to \cref{eq:fusion_mean} in the main text, and `nearest' to \cref{eq:fusion_nearest}.
For each baseline model, we compute the mean evaluation loss, $\Ltwo(\text{Mean})$, compared to the ground truth trajectories (over all seeds).
For stochastic models, for each evaluation sequence we also take $6$ samples of imputed trajectories, and also report the minimum evaluation loss, $\Ltwo(\text{Min.})$, over all samples, averaged over all seeds.
}
\begin{tabular}{P{3.95cm}
P{1.1cm}
P{1.6cm}
P{1.4cm}
c
S[table-format=1.3]@{\,\( \pm \)\,}
S[table-format=1.3]
S[table-format=1.3]@{\,\( \pm \)\,}
S[table-format=1.3]}
\toprule 
Model & Football-specific & Skip connection & Next-step cond. decoder & Bidirectional fusion mode & \multicolumn{2}{c}{$\Ls_{2}(\text{Mean})$} & \multicolumn{2}{c}{$\Ls_{2}(\text{Min.})$}\\
 \midrule
                  Role-invariant RNN &   \cmark &     $-$ &     $-$ &           $-$ &           0.940 &            0.01 &             \multicolumn{2}{c}{$-$} \\
           Bidir. Role-invariant RNN &   \cmark &     $-$ &     $-$ &         Mean &           0.442 &           0.004 &             \multicolumn{2}{c}{$-$} \\
           Bidir. Role-invariant RNN &   \cmark &     $-$ &     $-$ &  Nearest &           0.164 &           0.004 &             \multicolumn{2}{c}{$-$} \\
                 Role-invariant VRNN &   \cmark &  \xmark &     $-$ &           $-$ &           2.020 &            2.03 &           1.960 &           2.063 \\
                 Role-invariant VRNN &   \cmark &   \cmark &     $-$ &           $-$ &           0.958 &           0.009 &           0.953 &           0.009 \\
          Bidir. Role-invariant VRNN &   \cmark &  \xmark &     $-$ &         Mean &           0.510 &            0.02 &           0.486 &           0.019 \\
          Bidir. Role-invariant VRNN &   \cmark &  \xmark &     $-$ &  Nearest &           0.174 &           0.002 &           0.160 &           0.002 \\
          Bidir. Role-invariant VRNN &   \cmark &   \cmark &     $-$ &         Mean &           0.456 &           0.009 &           0.455 &           0.008 \\
          Bidir. Role-invariant VRNN &   \cmark &   \cmark &     $-$ &  Nearest &           0.167 &           0.002 &           0.166 &           0.002 \\
                              \midrule Linear &  \xmark &     $-$ &     $-$ &           $-$ &           0.658 &           0.081 &             \multicolumn{2}{c}{$-$} \\
                                LSTM &  \xmark &     $-$ &     $-$ &           $-$ &           1.579 &           0.019 &             \multicolumn{2}{c}{$-$} \\
                         Bidir. LSTM &  \xmark &     $-$ &     $-$ &         Mean &           0.751 &           0.009 &             \multicolumn{2}{c}{$-$} \\
                         Bidir. LSTM &  \xmark &     $-$ &     $-$ &  Nearest &           0.350 &           0.006 &             \multicolumn{2}{c}{$-$} \\
 Social LSTM~\citep{Alahi_2016_CVPR} &  \xmark &     $-$ &     $-$ &           $-$ &           1.049 &           0.274 &             \multicolumn{2}{c}{$-$} \\
                  Bidir. Social LSTM &  \xmark &     $-$ &     $-$ &         Mean &           0.457 &           0.011 &             \multicolumn{2}{c}{$-$} \\
                  Bidir. Social LSTM &  \xmark &     $-$ &     $-$ &  Nearest &           0.198 &           0.052 &             \multicolumn{2}{c}{$-$} \\
        GVRNN~\citep{yeh2019diverse} &  \xmark &  \xmark &  \xmark &           $-$ &           2.243 &           0.136 &           1.453 &           0.073 \\
        GVRNN~\citep{yeh2019diverse} &  \xmark &  \xmark &   \cmark &           $-$ &           2.447 &           1.197 &           2.400 &           1.231 \\
        GVRNN~\citep{yeh2019diverse} &  \xmark &   \cmark &  \xmark &           $-$ &           0.882 &           0.009 &           0.874 &           0.009 \\
        GVRNN~\citep{yeh2019diverse} &  \xmark &   \cmark &   \cmark &           $-$ &           0.865 &           0.018 &           0.852 &           0.017 \\
                Graph Imputer (Ours) &  \xmark &  \xmark &  \xmark &         Mean &           0.666 &           0.087 &           0.638 &            0.09 \\
                Graph Imputer (Ours) &  \xmark &  \xmark &  \xmark &  Nearest &           0.241 &            0.05 &           0.224 &           0.051 \\
                Graph Imputer (Ours) &  \xmark &  \xmark &   \cmark &         Mean &           1.106 &           0.368 &           1.094 &           0.381 \\
                Graph Imputer (Ours) &  \xmark &  \xmark &   \cmark &  Nearest &           0.404 &           0.102 &           0.397 &            0.11 \\
                Graph Imputer (Ours) &  \xmark &   \cmark &  \xmark &         Mean &           0.452 &           0.041 &           0.449 &            0.04 \\
                Graph Imputer (Ours) &  \xmark &   \cmark &  \xmark &  Nearest &           0.165 &           0.005 &           0.163 &           0.005 \\
                Graph Imputer (Ours) &  \xmark &   \cmark &   \cmark &         Mean &           0.418 &           0.005 &           0.414 &           0.005 \\
                Graph Imputer (Ours) &  \xmark &   \cmark &   \cmark &  Nearest &  \textbf{0.153} &  \textbf{0.003} &  \textbf{0.151} &  \textbf{0.003} \\
\bottomrule
\end{tabular}
\label{table:appendix_results}
\end{table}
\end{landscape}
\clearpage
}

\subsection{Additional result sweeps and baselines}

\Cref{table:appendix_results} presents additional comparative sweeps for the football off-screen player state estimation scenario.
In addition to the results in the main paper, this table includes the Role-invariant RNN baseline detailed in \cref{sec:appendix_baselines}.
The Role-invariant RNN model achieves quite similar performance as the Role-invariant VRNN counterpart, with the main distinction being that the former model is deterministic, in contrast to the latter;
in certain applications, the ability to resample the model (or, e.g., fine-tune the KL-regularization $\beta$ in \cref{eq:train_elbo} to increase or decrease the level of stochasticity in the model) can be quite useful from a practical perspective.

Additionally, \cref{table:appendix_results} includes sweeps over the bidirectional fusion modes \cref{eq:fusion_mean} and \cref{eq:fusion_nearest}.
For all bidirectional models, we observe that the nearest-observation weighted fusion mode \cref{eq:fusion_nearest} yields the lowest evaluation loss, primarily as it modulates the weighting of the directional updates (which deviate from the ground truth the longer they have not made an observation).

\subsection{Additional trajectory visualizations}

We provide a number of additional visualizations of trajectory predictions for the Graph Imputer and additional baselines in \cref{fig:traj_examples_supp_info_set1,fig:traj_examples_supp_info_set2,fig:traj_examples_supp_info_set3,fig:traj_examples_supp_info_set4}.
Here we also include a variant of the Graph Imputer which attains high trajectory sequence variance, which can be useful from a downstream analytical perspective when higher sample stochasticity is desired.

\begin{figure}[t]
    \setcounter{row}{1}%
    \setcounter{subfigure}{0}%
    \centering
    \begin{tabularx}{\textwidth}{p{0.08in}YYY}
        & \hspace{15pt} Example 1 & \hspace{15pt} Example 2 & \hspace{15pt} Example 3\\
        \resultsRow{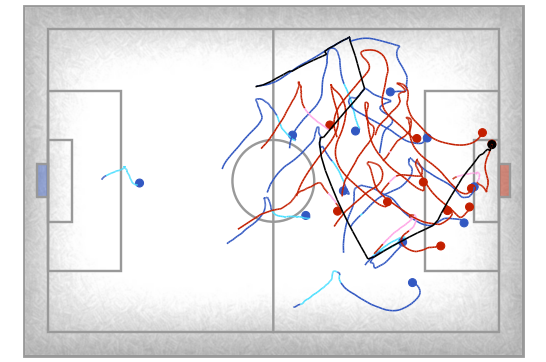}{Ground Truth}{0}
        \resultsRow{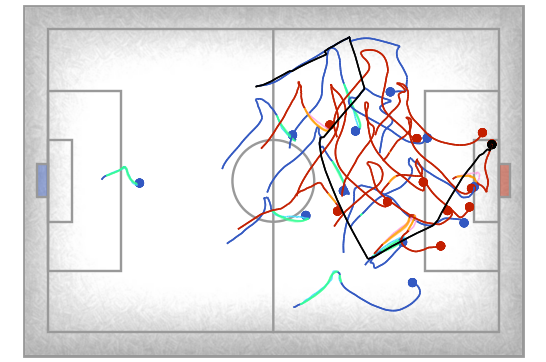}{Role-invariant VRNN}{0}
        \resultsRow{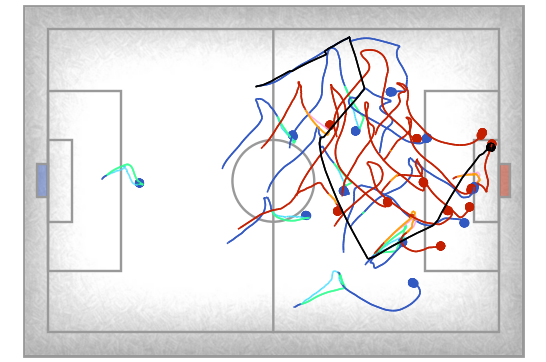}{Social LSTM}{0}
        \resultsRow{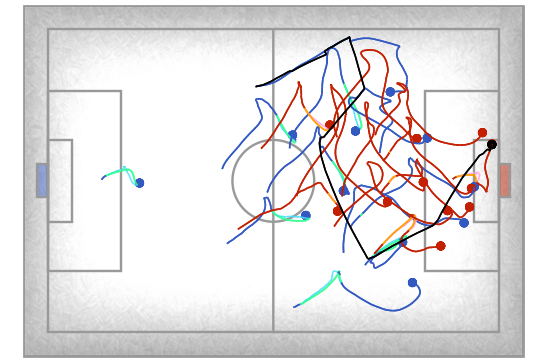}{Bidir. Social LSTM}{0}
        \resultsRow{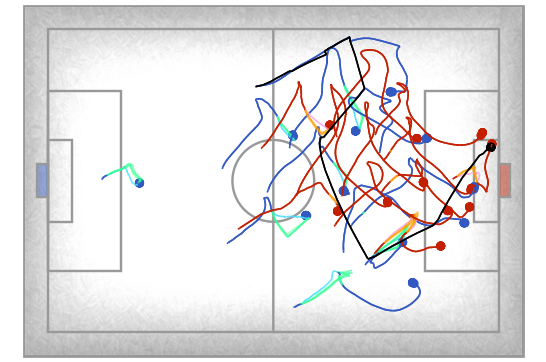}{GVRNN}{0}
        \resultsRow{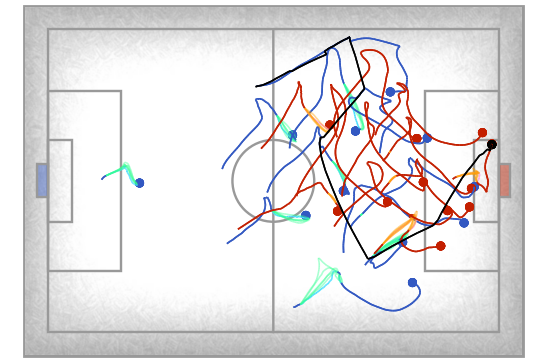}{\parbox{3cm}{\centering Graph Imputer\\(high var.)}}{0}
        \resultsRow{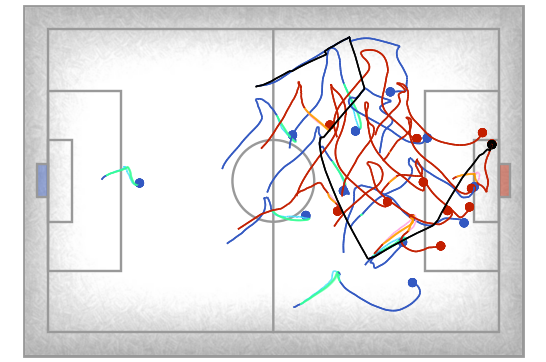}{Graph Imputer}{0}
    \end{tabularx}
    \hspace*{25pt}\includegraphics[trim={0 0 0 25pt}, width=0.95\textwidth]{figs/viz_legend.pdf}
    \caption{Trajectory visualizations (best viewed when zoomed in).
    Each column provides an example trajectory sequence, with the first row illustrating the ground truth, and subsequent rows showing results from various models, including the Graph Imputer (ours).
    }
    \label{fig:traj_examples_supp_info_set1}
\end{figure}

\begin{figure}[t]
    \setcounter{row}{1}%
    \setcounter{subfigure}{0}%
    \centering
    \begin{tabularx}{\textwidth}{p{0.08in}YYY}
        & \hspace{15pt} Example 4 & \hspace{15pt} Example 5 & \hspace{15pt} Example 6\\
        \resultsRow{trajs_supp_info/ground_truth_}{Ground Truth}{3}
        \resultsRow{trajs_supp_info/bidir_roleinvar_vrnn}{Role-invariant VRNN}{3}
        \resultsRow{trajs_supp_info/social_lstm}{Social LSTM}{3}
        \resultsRow{trajs_supp_info/bidir_social_lstm}{Bidir. Social LSTM}{3}
        \resultsRow{trajs_supp_info/gvrnn}{GVRNN}{3}
        \resultsRow{trajs_supp_info/graphimputer}{\parbox{3cm}{\centering Graph Imputer\\(high var.)}}{3}
        \resultsRow{trajs_supp_info/graphimputer_lowvar}{Graph Imputer}{3}
    \end{tabularx}
    \hspace*{25pt}\includegraphics[trim={0 0 0 25pt}, width=0.95\textwidth]{figs/viz_legend.pdf}
    \caption{Trajectory visualizations (best viewed when zoomed in).
    Each column provides an example trajectory sequence, with the first row illustrating the ground truth, and subsequent rows showing results from various models, including the Graph Imputer (ours).
    }
    \label{fig:traj_examples_supp_info_set2}
\end{figure}

\begin{figure}[t]
    \setcounter{row}{1}%
    \setcounter{subfigure}{0}%
    \centering
    \begin{tabularx}{\textwidth}{p{0.08in}YYY}
        & \hspace{15pt} Example 7 & \hspace{15pt} Example 8 & \hspace{15pt} Example 9\\
        \resultsRow{trajs_supp_info/ground_truth_}{Ground Truth}{6}
        \resultsRow{trajs_supp_info/bidir_roleinvar_vrnn}{Role-invariant VRNN}{6}
        \resultsRow{trajs_supp_info/social_lstm}{Social LSTM}{6}
        \resultsRow{trajs_supp_info/bidir_social_lstm}{Bidir. Social LSTM}{6}
        \resultsRow{trajs_supp_info/gvrnn}{GVRNN}{6}
        \resultsRow{trajs_supp_info/graphimputer}{\parbox{3cm}{\centering Graph Imputer\\(high var.)}}{6}
        \resultsRow{trajs_supp_info/graphimputer_lowvar}{Graph Imputer}{6}
    \end{tabularx}
    \hspace*{25pt}\includegraphics[trim={0 0 0 25pt}, width=0.95\textwidth]{figs/viz_legend.pdf}
    \caption{Trajectory visualizations (best viewed when zoomed in).
    Each column provides an example trajectory sequence, with the first row illustrating the ground truth, and subsequent rows showing results from various models, including the Graph Imputer (ours).
    }
    \label{fig:traj_examples_supp_info_set3}
\end{figure}

\begin{figure}[t]
    \setcounter{row}{1}%
    \setcounter{subfigure}{0}%
    \centering
    \begin{tabularx}{\textwidth}{p{0.08in}YYY}
        & \hspace{15pt} Example 10 & \hspace{15pt} Example 11 & \hspace{15pt} Example 12\\
        \resultsRow{trajs_supp_info/ground_truth_}{Ground Truth}{9}
        \resultsRow{trajs_supp_info/bidir_roleinvar_vrnn}{Role-invariant VRNN}{9}
        \resultsRow{trajs_supp_info/social_lstm}{Social LSTM}{9}
        \resultsRow{trajs_supp_info/bidir_social_lstm}{Bidir. Social LSTM}{9}
        \resultsRow{trajs_supp_info/gvrnn}{GVRNN}{9}
        \resultsRow{trajs_supp_info/graphimputer}{\parbox{3cm}{\centering Graph Imputer\\(high var.)}}{9}
        \resultsRow{trajs_supp_info/graphimputer_lowvar}{Graph Imputer}{9}
    \end{tabularx}
    \hspace*{25pt}\includegraphics[trim={0 0 0 25pt}, width=0.95\textwidth]{figs/viz_legend.pdf}
    \caption{Trajectory visualizations (best viewed when zoomed in).
    Each column provides an example trajectory sequence, with the first row illustrating the ground truth, and subsequent rows showing results from various models, including the Graph Imputer (ours).
    }
    \label{fig:traj_examples_supp_info_set4}
\end{figure}

\section{Cross-Sectional Overview of Closely Related Works}\label{sec:appendix_related_works}
\Cref{tab:table_references} provides an additional cross-section overview of the most closely related works to ours.
In this table, we summarize models that consider prediction of trajectories, detailing whether or not they are stochastic, consider the interactions of multiple agents in the system, target the imputation problem (as opposed to the typical forward-prediction setting), and use both forward- and backward-information.

\begin{table}[t]
\centering
\caption{\label{tab:table_references} Overview of the attributes of the different models covered in our related work section.
In this table, we summarize models that consider prediction of trajectories, detailing whether or not they are stochastic, consider the interactions of multiple agents in the system, target the imputation problem (as opposed to the typical forward-prediction setting), and use both forward- and backward-information.
}
\begin{tabular}{
p{3.0cm}
P{2.0cm}
P{1.2cm}
P{2.2cm}
P{1.2cm}
P{1.8cm}
}
    \toprule
      & Human trajectories & Stochastic & Considers  interactions between agents & Imputation & Forward \&  backward inf\\ 
     \midrule
     MBT \citep{hauri2020multimodal} & \cmark{} Basketball & \cmark & \xmark& \xmark& \xmark \\  
     \midrule
     VAIN \citep{hoshen2017} & \xmark{} No seq. pred. & \xmark & \cmark Interaction nets & \xmark& \xmark \\
     \midrule
     Imitation Learning \citep{le2017data, le2017coordinated} &  \cmark{} Football & \cmark & \cmark{} Conditioning & \xmark & \xmark\\
     \midrule
     Social LSTM \citep{Alahi_2016_CVPR} & \cmark{} Pedestrians &  \xmark & \cmark{} Social pooling layer & \xmark & \xmark\\
     \midrule
     Reciprocal learning nets~\citep{Sun_2020_CVPR} & \cmark{} Pedestrians  & \cmark  & \cmark{} Social pooling layer& \xmark & \cmark (As regularization)\\
     \midrule 
     EvolveGraph \citep{li2020Evolvegraph} &  \cmark{} Basketball & \cmark & \cmark & \xmark & \xmark \\
     \midrule
     GraphVRNN \citep{yeh2019diverse} & \cmark{} Basketball \& Football & \cmark & \cmark & \xmark & \xmark\\
      \midrule
     Volleyball trajectory prediction~\citep{Suda19} & \xmark{} Just the ball & \xmark & \xmark & \xmark & \xmark\\
     \midrule
     Time series classification~\citep{che2018recurrent} & \xmark & \xmark & \xmark & \cmark & \xmark \\
     \midrule
     GAN models~\citep{luo2018multivariate,luo2019e2gan}  & \xmark & \cmark & \cmark{} Conditioning & \cmark & \xmark\\
     \midrule
     BRITS~\citep{cao2018_brits} & \xmark & \xmark & \cmark{} Weighted conditioning & \cmark & \cmark\\
     \midrule
    M-RNN~\citep{DBLP:journals/corr/abs-1711-08742} & \xmark & \xmark & \xmark & \cmark & \cmark\\ 
    \midrule 
    Naomi~\citep{liu2019naomi} & \cmark{} Basketball & \cmark & \xmark & \cmark & \cmark\\
    \midrule 
    Baller2vec~\citep{alcorn2021baller2vec} & \cmark{} Basketball & \cmark & \cmark & \xmark & \xmark\\
    \midrule 
    Baller2vec++~\citep{alcorn2021b} & \cmark{} Basketball & \cmark & \cmark & \xmark & \cmark\\
    \bottomrule
\end{tabular}
\end{table}

\end{appendices}
\ifarXiv \else \end{bibunit} \fi

\end{document}